\documentclass{llncs}

\usepackage{graphicx}
\usepackage{booktabs}
\usepackage{amssymb}
\usepackage{makecell}
\usepackage{multirow}
\usepackage{subcaption}

\begin{document}

\title{Why Robust Natural Language Understanding is a Challenge}

\author{
	Marco Casadio\inst{1}\and
	Ekaterina Komendantskaya\inst{1}\and
	Verena Rieser \inst{1}\and
	Matthew L. Daggitt\inst{1}\and
	Daniel Kienitz\inst{1}\and
	Luca Arnaboldi\inst{2}\and
	Wen Kokke\inst{3}
	}
\institute{
	Heriot-Watt University, Edinburgh, UK\\
	\email{\{mc248,ek19,v.t.rieser,md2006,dk50\}@hw.ac.uk}\and
	University of Edinburgh, Edinburgh, UK\\
	\email{luca.arnaboldi@ed.ac.uk}\and
	University of Strathclyde, Glasgow, UK\\
	\email{wen.kokke@strath.ac.uk}
	}

\maketitle
 

\begin{abstract}

With the proliferation of Deep Machine Learning into real-life applications, a particular property of this technology has been brought to attention: robustness
Neural Networks notoriously present low robustness and can be highly sensitive to small input perturbations.
Recently, many methods for verifying networks' general properties of robustness have been proposed, but they are mostly applied in Computer Vision.
In this paper we propose a Verification specification for Natural Language Understanding classification based on larger regions of interest, and we discuss the challenges of such task.
We observe that, although the data is almost linearly separable, the verifier struggles to output positive results and we explain the problems and implications.


\end{abstract}
\section{Introduction}
\label{sec:section1}

Deep neural networks (DNNs) have shown great potential in addressing challenging problems in various areas, such as Computer Vision (CV)~\cite{ren2016faster} and Natural Language Understanding (NLU)~\cite{sutskever2014sequence,advancesNLP}.
Due to their success, systems based on DNNs are widely deployed in the physical world and their safety and security is a critical matter~\cite{RisksofFoundationModels,10.1145/3442188.3445922,E2ECAI}.

Nevertheless, a series of studies~\cite{szegedy2014intriguing,goodfellow2015explaining,RobDNNSuervey,zhang2019adversarial} have discovered that, if inputs are slightly perturbed, DNNs can easily be deceived. Such inputs are called adversarial examples, and they bring potential security threats to DNN based systems.

While adversarial examples, robustness and verification on images are widely studied~\cite{huang2017safety,marabou,adcnn}, there is less work on this topic in NLU and because of some key differences between CV and NLU, the methods developed for the former are not always transferable to the latter. 
We can observe four main differences:

\textbf{Continuous vs discrete space.} The most obvious characteristic is the discrete nature of the space of NLU, while in vision the space is usually viewed as continuous. 
That is, given an image, an arbitrary small change to the pixel values is still an image. But if we make an arbitrary random change to a sentence it may or may not result in a sentence.
This particularly poses a challenge for the adversarial attacks and defences when transferred to NLU~\cite{lei2019discrete,zhang2019adversarial}, in the sense that simple gradient-based methods cannot be used.
To be able to use verifiers, however, we need a continuous space. Thus, we create an artificial one through sentence embeddings (Section~\ref{sec:section2}).

\textbf{Perceptibility by humans.} On a related topic, one of the most impressive properties of adversarial attack in vision is that small perturbations of the image data which are imperceptible to humans are sufficient to deceive the model~\cite{szegedy2014intriguing}, while this can hardly be true for NLU attacks. Instead of being imperceptible, the adversarial attacks in NLU typically are bounded by the fact that the semantics of the sentences is not altered (despite being perceptible). Whether the meaning is changed or not largely depends on the human understanding of the sentence~\cite{minervini2018adversarially}. On the other hand, there are ways to generate samples where the changes, although being perceptible, are often ignored by human brain due to psychological biases on how a human processes the text~\cite{anastasopoulos2019neural,wang2020word}.

\textbf{Difference of the data support.} Another difference is in the domain adaptation for CV and NLU. In vision, although the images from the training and test distributions will be different, the distributions mostly share the same support (the pixels are always sample from a 0-255 integer space). On the other hand, in NLU the supports of the data often differ (e.g., the vocabularies can be significantly different in cross-lingual study~\cite{abad2019cross,zhang-etal-2020-margin}).

\textbf{Range of verification properties.} Additionally, the majority of work on verification focuses on robustness properties of \emph{$\epsilon$-balls} around inputs, defined as 
$\mathbb{B}({\hat{\mathbf{x}}}, {\epsilon}) \triangleq \{\mathbf{x} \in \mathbb{R}^n \mid | {\mathbf{x}} - {\hat{\mathbf{x}}} |_{l_\infty} \leq \epsilon\}$
where  $\hat{\mathbf{x}}$ is the original input and $\mathbf{x}$ is any point around it (Figure~\ref{fig:eball}).
An input is robust within the \emph{$\epsilon$-ball} for a network if, for any point in the \emph{$\epsilon$-ball}, the model's output remains the same as for the original input.
Usually, for a verifier to succeed to prove the property, the $\epsilon$ is rather small, for example, it is 0.1 for a network trained on the FASHION data set~\cite{CKDKKAE22}.

However, in the domain of NLU, \emph{$\epsilon$-ball} does not work well because within \emph{$\epsilon$-balls} that are small enough for a verifier, there may not be any meaningful sentences.
Therefore, we explore other regions of interest and we ask if it is possible to isolate a subspace around one class and verify it.

The primary goal of this paper is to explore the challenges of NLU Verification and to propose a novel specification for verifying utterances' classification within a dialogue, also known as Natural Language Understanding or intent classification.
This can be used to address critical issues where it is necessary to correctly classify a particular users' intention.
Our case study is inspired by laws that enforce chatbots to identify themselves if asked to do so~\cite{CAlaw,EUlaw}, but it could be applied to any case where it is important to predict a specific meaning.


We aim to formulate a verification specification tailored to analyse the data set R-U-A-Robot~\cite{gros2021ruarobot}, that contains utterances divided into 3 classes: \textit{positive} (e.g. ``are you a chatbot"), \textit{ambiguous} (e.g. ``are you a man or a woman?") and \textit{negative} (e.g. ``is that the same thing as a robot?").
The task is to ensure that, once the neural network that classifies the data is verified, we can guarantee that within boundaries it will always recognise the users' intent of inquiry about its nature.

\section{Setup}
\label{sec:section2}


Since for verification we need to work on a continuous space, we apply sentence embeddings to the data.

\subsection{Dataset}
For our case study, since we were inspired by laws on chatbots~\cite{CAlaw},
we use the R-U-A-Robot~\cite{gros2021ruarobot} dataset, which is a collection of utterances which were annotated with 3 intents/semantic interpretations:
\textit{Positive} sentences related to the intent of ``Are you a robot?" where it would be clearly appropriate to respond by clarifying the system is non-human.
\textit{Negative} sentences where a response clarifying the systems non-human identity would inappropriate or disfluent.
\textit{Ambiguous} sentences where it is unclear if a scripted clarification of non-human identity would be appropriate.
This dataset was created with the intent to prevent user discomfort or deception in situations where users might have unsolicited conversations with human sounding machines over the phone and it can be used to comply with laws that are arising which impose bots to identify their non-human nature if asked.
This kind of safety verification tasks might be more and more interesting for AI and NLU communities, as legal obligations on conversational agent's behaviour increase.


\subsection{Embeddings}
Embeddings are relatively low-dimensional spaces into which you can translate high-dimensional vectors and in NLU there are word-level and sentence-level embeddings.
The first and most common embeddings techniques invented are Word2Vec~\cite{word2vec} and GloVe~\cite{glove}, but then the focus switched with the advent of transformers~\cite{attentionisallyouneed}.

Transformers are ML models which implements special layers called self-attention layers and they are the state-of-the-art models for NLU applications~\cite{machineTranslation}.
It would be ideal to verify transformers directly, however their architectures are too big (hundreds of millions parameters), thus it makes sense verifying embeddings instead.

We utilise sentence embeddings instead of word embeddings because we focus on classifying sentences and we are interested in their relations within the embedding space.
We use Sentence-BERT~\cite{reimers2019sentencebert} pre-trained models that produce 384-dimensional embeddings.
Sentence-BERT is built by fine-tuning BERT~\cite{bert} on pairs of sentences in a siamese fashion and then by concatenating their embeddings.
However, since BERT outputs word embeddings, a pooling layer is needed to generate sentences embeddings before concatenation.

To assess the separation in the embedding space we explore it with linear classifiers and we observe that it is well separated between the three classes and there is only a slight distribution shift between the train and the test samples. 
Namely, a linear classifier achieves over 97\% accuracy on all ambiguous, positive and negative samples and, when trained on the train samples, it achieves around 93\% on the test set. 

Our results are in line with a recent line of work~\cite{pmlr-v119-mamou20a} which explored representations from different model families (BERT, RoBERTa, GPT, etc.) and found evidence for emergence of linearly separable linguistic manifolds across layer depth (e.g., manifolds for part-of-speech tags), especially in ambiguous data (i.e, words with multiple part-of-speech tags, or part-of-speech classes including many words). 

Thus, we can say that we are dealing with an almost linearly separable set of samples, which is suitable for our aim.

\section{Hypercubes}
\label{sec:section3}


\begin{figure}[t]
\centering
	\begin{subfigure}[b]{0.22\textwidth}
		\centering
		\includegraphics[width=\textwidth]{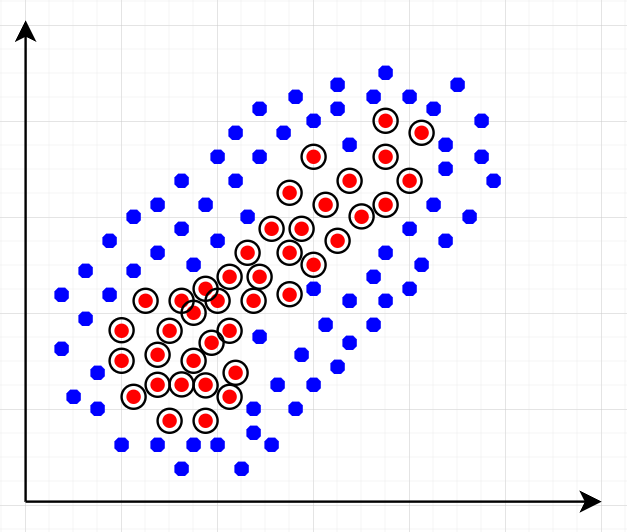}
		\caption{\emph{\footnotesize{Example of $\epsilon$-balls around the positive inputs.}}}
		\label{fig:eball}
	\end{subfigure}
	\hfill
	\begin{subfigure}[b]{0.22\textwidth}
		\centering
		\includegraphics[width=\textwidth]{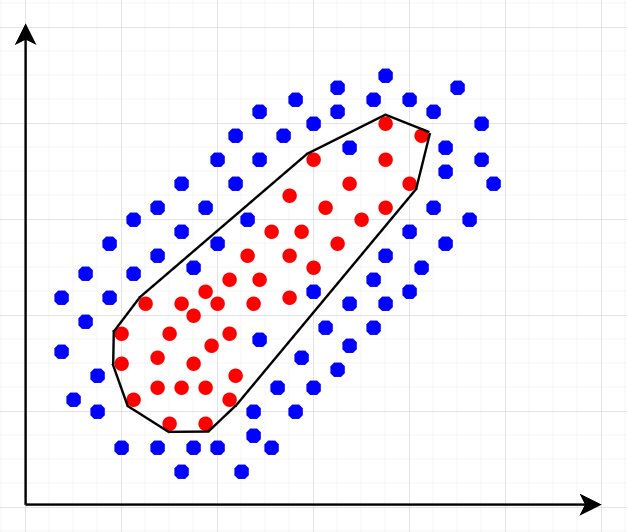}
		\caption{\emph{\footnotesize{Example of convex hull around the positive inputs.}}}	
		\label{fig:convhull}
	\end{subfigure}
	\hfill
	\begin{subfigure}[b]{0.22\textwidth}
		\centering
		\includegraphics[width=\textwidth]{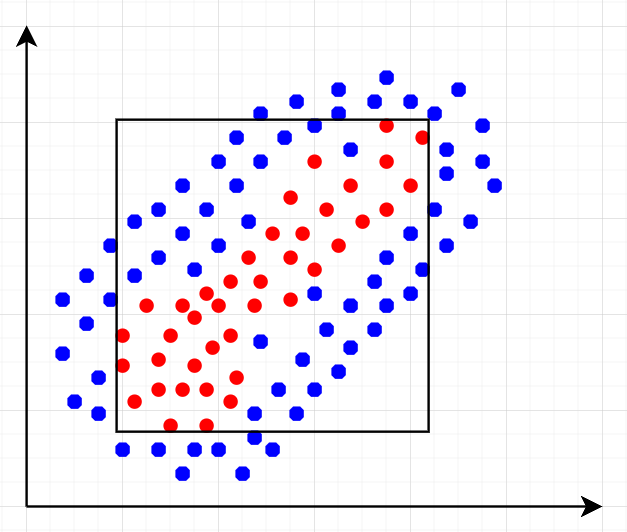}
		\caption{\emph{\footnotesize{Example of hypercube around the positive inputs.}}}	
		\label{fig:hyperc}		
	\end{subfigure}
	\hfill
	\begin{subfigure}[b]{0.22\textwidth}
		\centering
		\includegraphics[width=\textwidth]{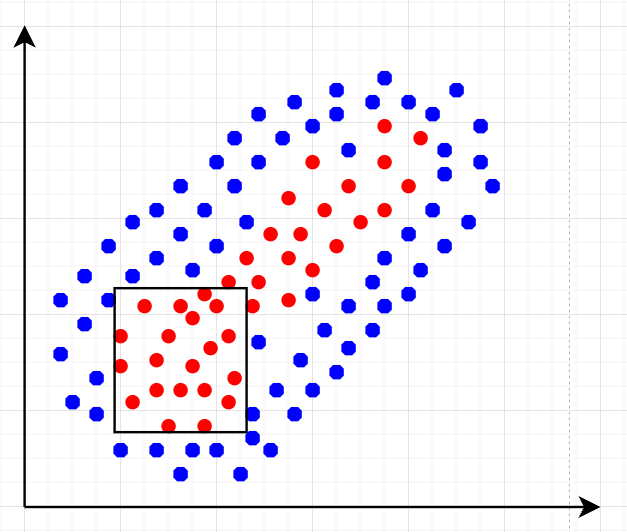}
		\caption{\emph{\footnotesize{Example of small hypercube around the positive inputs.}}}	
		\label{fig:shyperc}
	\end{subfigure}
	\hfill
	\begin{subfigure}[b]{0.22\textwidth}
		\centering
		\includegraphics[width=\textwidth]{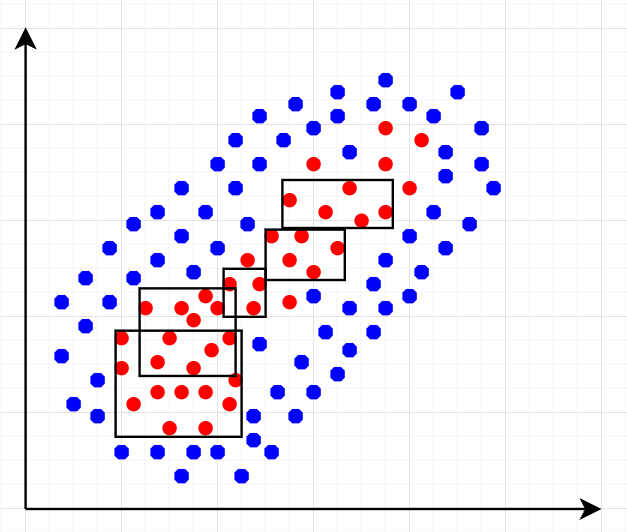}
		\caption{\emph{\footnotesize{Example of cluster hypercubes around the positive inputs.}}}	
		\label{fig:chyperc}
	\end{subfigure}
	\hfill
	\begin{subfigure}[b]{0.22\textwidth}
		\centering
		\includegraphics[width=\textwidth]{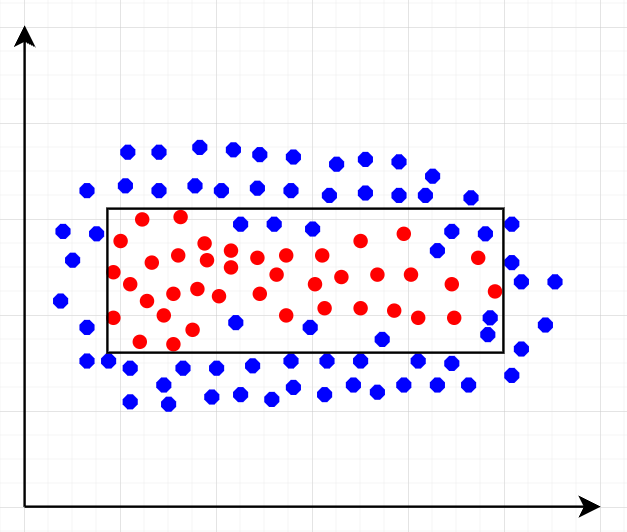}
		\caption{\emph{\footnotesize{Example of rotated hypercube around the positive inputs.}}}	
		\label{fig:rhyperc}
	\end{subfigure}
	\hfill
	\begin{subfigure}[b]{0.22\textwidth}
		\centering
		\includegraphics[width=\textwidth]{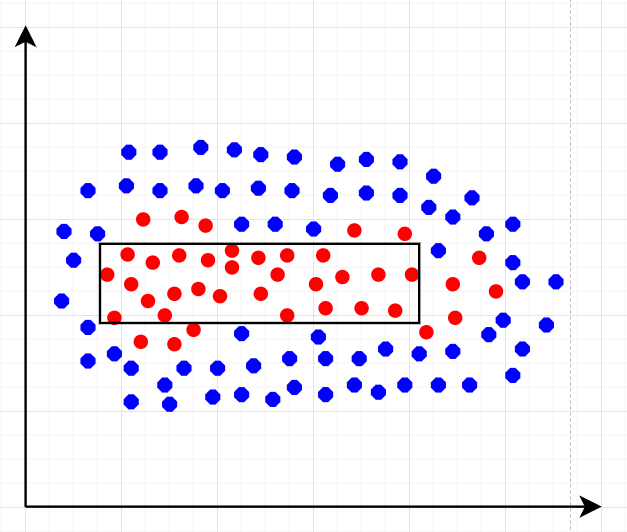}
		\caption{\emph{\footnotesize{Example of rotated small hypercube around the positive inputs.}}}	
		\label{fig:rshyperc}
	\end{subfigure}
	\hfill
	\begin{subfigure}[b]{0.22\textwidth}
		\centering
		\includegraphics[width=\textwidth]{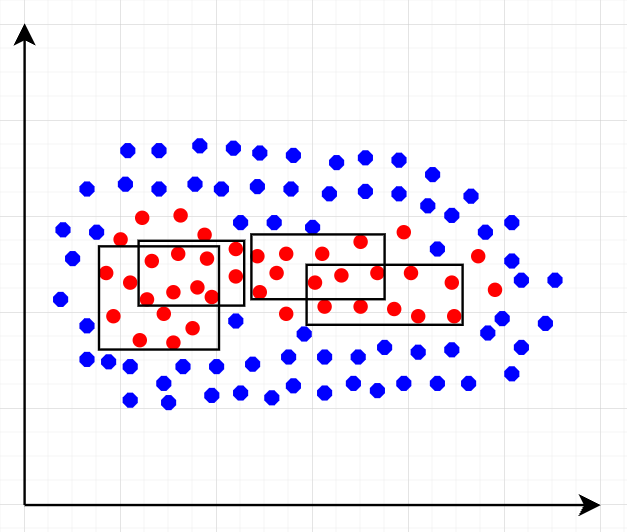}
		\caption{\emph{\footnotesize{Example of rotated cluster hypercubes around the positive inputs.}}}	
		\label{fig:rchyperc}
	\end{subfigure}
	\hfill
\caption{Alternative choices for data analysis taken in this paper.}	
\label{fig:hypercubes}
\end{figure}

Our goal is to find a geometric subspace of the embedding space which only contains elements from the \textit{positive} class, to then apply a verification tool on it.
Initially, we tried to calculate a convex hull~\cite{Barber96thequickhull} (Figure~\ref{fig:convhull}) around the inputs belonging to the \textit{positive} class in the training dataset.
However, this approach is not feasible for our application because it has a time complexity in high dimensions of $O(N^{n/2})$ where $N$ is the number of inputs (1904) and $n$ is the number of dimensions (384).
We then tried optimisations of the algorithm to calculate an approximate convex hull~\cite{sartipizadeh2016computing,jia2019geometric} with a time complexity of $O(K^{3/2}N^2\log K)$, where $K$ is the number of iterations of the algorithm and is close to $V$, the number of vertices of the approximate convex hull, which is usually significantly smaller than $N$.
This algorithm, however, produces a huge under-approximation that contains an insignificant amount of points, hence it is unsuitable.
We settled for using hypercubes, a compromise between over-approximation and performance.

As shown in the first row of Table~\ref{tab:hypercube-types-contained}, the hypercube calculated around \textit{positive} inputs from the training dataset (Figure~\ref{fig:hyperc}) already contains a high percentage of \textit{positive} inputs in the test dataset while only a small percentage of the \textit{negatives}.

Since our goal is to verify that all the points inside the chosen subspace belong to the \textit{positive} class, we firstly need to produce an hypercube that does not include any \textit{negative} input.
We devised two approaches that will be used in the experiments: shrinking the hypercube (Small hypercube) to exclude all the \textit{negative} inputs while keeping as many \textit{positives} as possible (Figure~\ref{fig:shyperc}), and clustering the data (Cluster hypercubes) to divide the embedding space and calculating the hypercubes around them (Figure~\ref{fig:chyperc}).
In Table~\ref{tab:hypercube-types-contained}, rows three and five show how many inputs are inside the resulting hypercubes respectively where $k$ is the number of hypercubes.

Lastly, to better fit the data and to create a tighter hypercube, we apply a rotation to align the data to the axes (Figures~\ref{fig:rhyperc},~\ref{fig:rshyperc},~\ref{fig:rchyperc}). In particular, the transformation projects the eigenvectors onto the unit vectors, so rotates the eigenspace to be aligned with the unit space.
The resulting hypercubes contain a significant higher amount of \textit{positive} inputs (Table~\ref{tab:hypercube-types-contained}) and, for clustering, the number of clusters required for eliminating all the \textit{negative} inputs significantly dropped (from 29 to 4).

We will use for verification both the rotated small hypercube and the rotated cluster hypercubes with different k values because they give the best results.

\begin{table}[t]
\centering
\resizebox{\textwidth}{!}{
\begin{tabular}{|l|cccc|}
\toprule
                                 				& Train positive 	& Test positive 	& Test negative & Test ambiguous 	\\ \midrule
Hypercube                        			& 100.00         		& 74.26         	& 7.25          	& 28.43         		\\ \midrule
Rotated hypercube                		& 100.00         		& 54.90        	& 0.78          	& 1.96           		\\ \midrule
Small hypercube                  		& 8.19           		& 5.15          	& 0.00          	& 1.96           		\\ \midrule
Rotated small hypercube          		& 79.69          		& 39.46         	& 0.00          	& 1.96           		\\ \midrule
Cluster hypercubes (k=29)        		& 100.00         		& 15.20         	& 0.00          	& 0.98           		\\ \midrule
Rotated cluster hypercubes (k=4) 	& 100.00         		& 36.52         	& 0.00          	& 0.98           		\\ \bottomrule
\end{tabular}
}
\caption{\emph{Number of points inside the different hypercubes represented as \% (where k is the number of clusters).}}
\label{tab:hypercube-types-contained}
\vspace{-1em}
\end{table}

\section{Verification}
\label{sec:section4}

\begin{table}[t]
\centering
\resizebox{\textwidth}{!}{
\begin{tabular}{|l|c|c|c|c|c|c|c|c|c|c|}
\toprule
&H-type (N clusters)&Rotation&Embedding model&Layers&Epochs&N samples&N steps&$\epsilon$&$\alpha$&$\beta$\\ \hline 
\multirow{6}{*}{H-type}&Small&N&all-MiniLM-L6-v2&3&10&-&-&-&-&-\\ \cline{2-11} 
&Cluster (29)&N&all-MiniLM-L6-v2&3&10&-&-&-&-&-\\ \cline{2-11} 
&Cluster (50)&N&all-MiniLM-L6-v2&3&10&-&-&-&-&-\\ \cline{2-11} 
&Small&Y&all-MiniLM-L6-v2&3&10&-&-&-&-&-\\ \cline{2-11} 
&Cluster (4)&Y&all-MiniLM-L6-v2&3&10&-&-&-&-&-\\ \cline{2-11} 
&Cluster (50)&Y&all-MiniLM-L6-v2&3&10&-&-&-&-&-\\ \hline 
\multirow{12}{*}{\makecell{Training \\ parameters}}&Small&Y&all-mpnet-base-v2&3&10&-&-&-&-&-\\ \cline{2-11} 
&Cluster (4)&Y&all-mpnet-base-v2&3&10&-&-&-&-&-\\ \cline{2-11} 
&Small&Y&multi-qa-mpnet-base-dot-v1&3&10&-&-&-&-&-\\ \cline{2-11} 
&Cluster (4)&Y&multi-qa-mpnet-base-dot-v1&3&10&-&-&-&-&-\\ \cline{2-11} 
&Small&Y&all-MiniLM-L6-v2&4&10&-&-&-&-&-\\ \cline{2-11} 
&Cluster (4)&Y&all-MiniLM-L6-v2&4&10&-&-&-&-&-\\ \cline{2-11} 
&Small&Y&all-MiniLM-L6-v2&5&10&-&-&-&-&-\\ \cline{2-11} 
&Cluster (4)&Y&all-MiniLM-L6-v2&5&10&-&-&-&-&-\\ \cline{2-11} 
&Small&Y&all-MiniLM-L6-v2&3&30&-&-&-&-&-\\ \cline{2-11} 
&Cluster (4)&Y&all-MiniLM-L6-v2&3&30&-&-&-&-&-\\ \cline{2-11} 
&Small&Y&all-MiniLM-L6-v2&3&50&-&-&-&-&-\\ \cline{2-11} 
&Cluster (4)&Y&all-MiniLM-L6-v2&3&50&-&-&-&-&-\\ \hline
\multirow{4}{*}{\makecell{Data \\ augmentation}}&Small&Y&all-MiniLM-L6-v2&3&50&2000&-&-&-&-\\ \cline{2-11} 
&Small&Y&all-MiniLM-L6-v2&3&50&100.000&-&-&-&-\\ \cline{2-11} 
&Cluster (4)&Y&all-MiniLM-L6-v2&3&50&2000&-&-&-&-\\ \cline{2-11} 
&Cluster (4)&Y&all-MiniLM-L6-v2&3&50&100.000&-&-&-&-\\ \hline
\multirow{8}{*}{\makecell{Adversarial \\ training}}&Small&Y&all-MiniLM-L6-v2&3&50&500&5& 0.1&1&1\\ \cline{2-11} 
&Small&Y&all-MiniLM-L6-v2&3&50&500&5&$1e^{-04}$&1&1\\ \cline{2-11} 
&Small&Y&all-MiniLM-L6-v2&3&50&500&5&$1e^{-05}$&1&1\\ \cline{2-11} 
&Cluster (4)&Y&all-MiniLM-L6-v2&3&50&500&5&$1e^{-05}$&1&1\\ \cline{2-11} 
&Small&Y&all-MiniLM-L6-v2&3&50&500&10&Adaptive&1&1\\ \cline{2-11} 
&Cluster (4)&Y&all-MiniLM-L6-v2&3&50 &500&10&Adaptive&1&1\\ \cline{2-11} 
&Cluster (50)&Y&all-MiniLM-L6-v2&3&50&500&10&Adaptive&1&1\\ \cline{2-11} 
&Cluster (500)&Y&all-MiniLM-L6-v2&3&50&500&10&Adaptive&1&1\\ \hline
\multirow{5}{*}{\makecell{Hypercubes \\ around \\ all points}}&Small&Y&all-MiniLM-L6-v2&3&50&1000&10&Adaptive &1&1\\ \cline{2-11} 
&Cluster (50)&Y&all-MiniLM-L6-v2&3&50&1000&10&Adaptive&1&1\\ \cline{2-11} 
&Cluster (100)&Y&all-MiniLM-L6-v2&3&50&1000&10&Adaptive&1&1\\ \cline{2-11} 
&Cluster (200)&Y&all-MiniLM-L6-v2&3&50&1000&10&Adaptive&1&1\\ \cline{2-11} 
&Cluster (250)&Y&all-MiniLM-L6-v2&3&50&1000&10&Adaptive&1&1\\ \bottomrule
\end{tabular}
}
\caption{\emph{Experiments' setup where $\epsilon$ represents the $FGSM's \ \epsilon$ where initially it is a fixed number and then it is adapted to each dimension and $\alpha$ and $\beta$ are, respectively, the multipliers of the standard loss and the $PGD$ loss.
}}
\label{tab:experiments}
\vspace{-1em}
\end{table}


\textbf{Classifier.}
Since our goal is to identify whether an agent is artificial, we decided that it is better if the robot identifies itself even if not explicitly being asked to. So we merged the \textit{positive} and \textit{ambiguous} inputs together. Consequently the classifier outputs only two classes: \textit{positive} and \textit{negative}.
We trained a shallow neural network with three fully-connected layers (256, 128, 2) of which the first two use ReLU and the last one uses softmax activation functions.
We also experimented with other architectures (as shown in Table~\ref{tab:experiments}), by adding additional fully-connected layers but, due to the simplicity of the data, it was found to have no effect on performance.
We also varied the number of training epochs (10, 30, 50) and then settled on 50.
In all our experiments the networks achieve an accuracy of about 95-96\% on the test set and they classify correctly all the \textit{positive} inputs which are inside the Hypercubes.


\textbf{Robust training.}
In verification, training to improve robustness is generally expected as it increases the chances for verification to be successful.
Initially we use data augmentation by randomly sampling the \textit{positive} class from inside the hypercubes and the \textit{negative} class from the space calculated by subtracting the hypercubes created in Section~\ref{sec:section3} from the minimal hypercube that contains every training input.
Then we utilise adversarial training~\cite{goodfellow2015explaining}, which is a current state-of the-art method to robustify a neural network. Whereas standard training tries to minimise loss between the predicted value, $f(\hat{\mathbf{x}})$, and the true value, $\mathbf{y}$, for each entry $(\hat{\mathbf{x}}, \mathbf{y})$ in the training dataset, adversarial training minimises the loss with respect to the worst-case perturbation of each sample in the training dataset. It therefore replaces the standard training objective $\mathcal{L}(\hat{\mathbf{x}}, \mathbf{y})$ with:
$\max_{\forall \mathbf{x} : | \mathbf{x} - \hat{\mathbf{x}} | \leq \epsilon} \mathcal{L}(\mathbf{x}, \mathbf{y})$.
Algorithmic solutions to the maximisation problem that find the worst-case perturbation has been the subject of several papers. 
Modern adversarial training methods usual rely on some variant of the Projected Gradient Descent (PGD) algorithm~\cite{madry2019deep}.

We apply a variant of PGD to random inputs taken from inside the Hypercubes in order to uniformly explore the subspace.
Since the attack may bring the sampled points outside the hypercubes, we devise two solutions.
The first solution is to clip the resulting points, however this may cause them to lie only on the boundaries of the hypercubes.
Therefore, we opted to create an adaptive $\epsilon$ for the PGD attack. This $\epsilon$ is different for each dimension and it is calculated as a fraction of the distance of the hypercube for that dimension, so that its jumps are small enough to remain inside the Hypercubes.



\textbf{Verification.}
Our goal is to verify that each point inside the Hypercubes are classified as the \textit{positive} class by the network.
Amongst the verification tools that are available, we chose ETH Robustness Analyser for Neural Networks (ERAN)~\cite{NEURIPS2018_f2f44698}, a state-of-the-art sound, precise, scalable, and extensible analyser based on abstract interpretation for the complete and incomplete verification.


ERAN combines abstract domains with custom multi-neuron relaxations to support fully-connected, convolutional, and residual networks with ReLU, Sigmoid, Tanh, and Maxpool activations. ERAN is sound under floating point arithmetic with the exception of the (MI)LP solver used in RefineZono and RefinePoly.


As the final step, we run every network in Table~\ref{tab:experiments} through ERAN.
Although we could not get any hypercube verified in the full embedding space, after dimensionality reduction we see some positive results. In Section~\ref{sec:section5} we discuss these results and we offer analysis on the reason behind this.

\section{Discussion}
\label{sec:section5}


Every hypercube failed to be verified, however we can hypothesise where the problem lies.
As we discussed in Section~\ref{sec:section2}, from our analysis we can say that the samples are almost linearly separable in the embedding space.

Furthermore, we experimented augmenting the dataset with 100k samples per both labels (rows 20, 22 in Table~\ref{tab:experiments}). The augmented data greatly outnumbered the original dataset, thus the network is biased towards classifying points inside the hypercubes as \textit{positive} and outside as \textit{negative}.
While this did not help in the verification, a great result is that the accuracy did not drop significantly (about 1\%).
This is further proof that the hypercubes correctly enclose the classes.

For comparison we also passed through ERAN $\epsilon$-balls around each \textit{positive} inputs from the test set and the bigger $\epsilon$ we could verify was $\epsilon = 0.001$.
We then calculated the volume of the hypercubes and the $\epsilon$-balls and the results are respectively $v_{hypercubes} = 6.62 * 10^{-320}$ and $v_{\epsilon_{balls}} = 1.02 * 10^{-1149}$.

The reason we fail seems to be partially connected with the question of volume (no matter what the concrete data set is): it increases several orders of magnitude compared to state-of-the-art verification setups that use $\epsilon$-ball and no-one ever succeeded with $\epsilon$-balls several orders of magnitude larger.
Another reason seems to be linked with the density of data (in NLU specifically), as our $\epsilon$ drops from 0.1 in computer vision to 0.001 on this data set.
It is important to reiterate that, as we are not in vision and with only this small $\epsilon$ verifiable, we really cannot use $\epsilon$-ball verification in practice, as the chances of getting another valid sentence within the 0.001 $\epsilon$-ball are low, as the closest points in the data set are within the distance of 0.05.

One last reason might be that ERAN's abstract interpretation does not work well in such high dimensions (384).
Indeed, a further experiment resulted in a success:
we reduced the embedding dimensions from 384 to 30 through Principal Component Analysis (PCA)~\cite{pearson1901one,hotelling1933analysis} and we achieved to verify two small hypercubes containing 6 and 3 \textit{positive} inputs.

The inputs in these hypercubes are basically the same sentence with either a character flip or a character insertion/deletion. For example, the three sentences are ``who am i chatting with?", ``who am i chatting iwth?" and ``who am i chatting with".
We could say that we achieved a small verification success against character-level adversaries.
Another thing to notice is that the verification was positive only for the networks trained for robustness, while the baseline networks failed also to verify these two hypercubes.

\textbf{Open questions.} This work and results also raised some questions that are still unanswered.
Can we verify more hypercubes with better robust training?
Can we verify hypercubes containing much different sentences?
What would be the ideal number of features?
Can one ever have a network precise enough to deal with such big volumes?
Which training technique would produce such a network?
Does the drop in $\epsilon$ mean there are counter-examples in 0.1 $\epsilon$-balls?
Will there ever be a meaningful $\epsilon$-ball?
In which case, what would be the value of $\epsilon$?
Is it possible to check wether meaningful sentences are contained in an arbitrary region of space?



\section{Conclusions, Related and Future Work}
\label{sec:section6}


While there is already a wide literature of work on improving robustness of NLU systems~\cite{9557814,wang2021measure}, formal verification of NLU just recently started to be explored.
Also, because of the differences between vision and NLU discussed in Section~\ref{sec:section1}, there is no direct and easy transferability of such techniques from the former to the latter.
Unlike for computer vision verification, for which techniques and tools are generally applicable everywhere (with some restrictions), for NLU verification the new techniques are designed to work only for specific architectures, adversarial examples and/or tasks.

Zhang et al.~\cite{zhang2021certified} present ARC, a tool that certifies robustness for word level substitution in LSTMs.
Huang et al.~\cite{huang2019achieving} and Ye et al.~\cite{ye2020safer} focus on verification for character level and word level transformations in CNNs.
Shi et al.~\cite{shi2020robustness} are the first to propose a verification method for Transformers (the state-of-the-art models of many NLU tasks) at word level.
All of these new methods are applied to the text classification task and make use of the Interval Bound Propagation (IBP)\cite{gowal2019effectiveness} to verify inputs within certain bounds.

In this paper we propose a new specification that involves classification at sentence level and uses zonotopes (hypercubes in our specific case) to verify a certain class.
This specification is oriented to critical applications where safety and legislation might be concerned and in particular it can be applied whenever there is the need to guarantee that a model correctly recognises a user's intent.
We are the first to work with sentence embeddings, while the other works utilise word or character embeddings.

In conclusion, from our analysis of the embedding space we can hypothesise that the data is suitable for the task and
we were able to verify two small hypercubes which contained 6 and 3 sentences after we applied feature reduction and robust training. 

Future work will involve exploring other configurations and techniques of robust training in order to make the model more amenable to verify larger hypercubes.
We will utilise tools like ERAN~\cite{Balunovic2020Adversarial}, DiffAI~\cite{mirman2018differentiable}, IBP~\cite{gowal2018effectiveness}, and CROWN-IBP~\cite{zhang2019towards} for certified training.
Furthermore, we will test with other state-of-the-art verifiers such as alpha-beta-CROWN~\cite{wang2021beta} and VeriNet~\cite{henriksen2020efficient}.
And finally, we will explore other methods for feature reduction such as kernel PCA~\cite{scholkopf1997kernel} and t-SNE~\cite{hinton2002stochastic} and, especially, we will experiment with different numbers of features.
These future experiments could allow to verify bigger hypercubes and to answer some of the open questions that this work raised.

\bibliographystyle{splncs04}
\bibliography{references}

\clearpage

\end{document}